\documentclass[manuscript,screen,nonacm]{acmart}
\usepackage{caption}
\usepackage{subcaption}
\usepackage{graphicx}
\usepackage{color}
\usepackage{placeins}

\usepackage{breakurl}
\usepackage{lipsum}
\usepackage[ruled,vlined]{algorithm2e}
\usepackage{scalerel}
\usepackage{hhline}
\usepackage{tikz}
\usepackage{adjustbox}
\usepackage[outdir=./]{epstopdf}

\AtBeginDocument{%
  \providecommand\BibTeX{{%
    \normalfont B\kern-0.5em{\scshape i\kern-0.25em b}\kern-0.8em\TeX}}}

\setcopyright{acmcopyright}
\copyrightyear{2022}
\acmYear{2022}
\acmDOI{XXXXXXX.XXXXXXX}

\acmPrice{15.00}
\acmISBN{978-1-4503-XXXX-X/18/06}

\begin{document}

\title{Verification system based on long-range iris and Graph Siamese Neural Networks}

\author{Francesco Zola}
\authornote{F. Zola is also with the Public University of Navarre and the Netherlands Forensic Institute.}
\email{fzola@vicomtech.com}
\orcid{0000-0002-1733-5515}
\author{Jose Alvaro Fernandez-Carrasco}
\email{jafernandez@vicomtech.org}
\orcid{0000-0001-7574-7123}
\author{Jan Lukas Bruse}
\orcid{0000-0002-5774-1593}
\email{jbruse@vicomtech.org}
\affiliation{%
  \institution{Vicomtech Foundation, Basque Research and Technology Alliance}
  \country{Donostia}
  \country{Spain}}

\author{Mikel Galar}
\email{mikel.galar@unavarra.es}
\orcid{0000-0003-2865-6549}
\affiliation{%
  \institution{Institute of Smart Cities, Department of Statistics, Computer Science and Mathematics, Public University of Navarre}
  \country{Pamplona}
  \country{Spain}}

\author{Zeno Geradts}
\orcid{0000-0001-5912-5295}
\email{z.geradts@nfi.nl}
\affiliation{%
  \institution{Netherlands Forensic Institute}
  \country{The Hague}
  \country{The Netherlands}
 }
 
\renewcommand{\shortauthors}{Zola, et al.}

\begin{abstract}
Biometric systems represent valid solutions in tasks like user authentication and verification, since they are able to analyze physical and behavioural features with high precision. However, especially when physical biometrics are used, as is the case of iris recognition, they require specific hardware such as retina scanners, sensors, or HD cameras to achieve relevant results. At the same time, they require the users to be very close to the camera to extract high-resolution information. For this reason, in this work, we propose a novel approach that uses long-range (LR) distance images for implementing an iris verification system. More specifically, we present a novel methodology for converting LR iris images into graphs and then use Graph Siamese Neural Networks (GSNN) to predict whether two graphs belong to the same person. In this study, we not only describe this methodology but also evaluate how the spectral components of these images can be used for improving the graph extraction and the final classification task. Results demonstrate the suitability of this approach, encouraging the community to explore graph application in biometric systems.
\end{abstract}




\maketitle

\section{Introduction}
Soft biometric information, i.e., physical and behavioural traits such as gender, height, iris, and voice, can be helpful in security, authentication, and validation processes \cite{reid2013soft}. The main advantage of using biometric information over traditional methods is that instead of requiring information that the user should know or possess (password, codes, PIN, etc.), they use characteristics that univocally and biologically define the users (fingerprints, iris, face, etc.). In particular, these characteristics are universal (all users can be measured), singular (each user has its own measures), permanent in time and context, and can be quantitatively measured \cite{Matyas-R:advantages}.

Soft biometrics can be divided into two groups: physical and behavioural biometrics. Techniques of the first category use physical characteristics like face, iris, and fingerprint for their tasks \cite{alsaadi2021study}, whereas techniques of the second one, use information extracted from user behaviours such as signature, voice, and keyboard typing \cite{oak2018literature}.

Among the physical biometrics, face \cite{ghalleb2013face} and fingerprint \cite{ailisto2006soft} methodology have been the most explored, and have already been used in many real-world applications such as airport scanners, banking, military access control, smartphones or forensics \cite{Muley-K:banking, Bud:faceid}. However, in the last decade, the use of iris has begun to attract interest in applications such as gender classification \cite{khan2021authentication}, iris liveness detection \cite{chen2018iris}, border control \cite{sequeira2018protect} and citizen confirmation \cite{jayanthi2021effective}. 

In fact, iris biometric represents a secure biometric with low forgery and error rates due to its highly certain features \cite{rui2018survey}. Furthermore, this biometric information is usually combined with Artificial Intelligence (AI) and Machine Learning techniques (ML) in order to implement user identification and verification systems. However, one of the main drawbacks of these iris-based approaches, is that for their real application, they require specific hardware such as retina scanners, sensors, HD cameras, etc, to achieve relevant results in security and authentication tasks. At the same time, they require the users to be very close to the camera to extract high-resolution information. Consequently, the majority of the works in the literature that propose to use long-range (LR) iris images are focused on improving the acquisitions system using specific technologies \cite{NGUYEN2017123, venugopalan2011long, de2010design, bashir2008eagle}.

In contrast, in this work, we present a novel approach for LR iris verification based on graph analysis. Our idea consists of extracting relevant information from each LR iris image captured from standard cameras and representing it as a graph. In fact, graph representations are widely applied in social media \cite{schall2014link}, biology \cite{abbas2021application}, program analysis \cite{yan2019classifying} and cybersecurity \cite{zola2022network} tasks, showing promising results. However, they are not extensively applied in biometrics tasks. For this reason, in this work, not only we propose to exploit their potential for representing LR iris images, but also to use these graphs for training a verification system based on graph Machine Learning. In particular, our idea is to implement Graph Siamese Neural Network (GSNN) for identifying the similarities and differences between two graphs, returning as a result whether the two graphs belong to the same person. The whole process is firstly validated using \textit{original} LR iris images, then using \textit{spectral enhanced} iris images, i.e., LR iris images in which convolutional filters are applied for highlighting spectral components. Finally, the best configurations are tested increasing the number of users to be distinguished, to evaluate the generalization of our approach. To the best of our knowledge, this is the first work that proposes to extract graph representations from LR iris images and use them directly for user verification using GSNN.

The rest of the paper is organized as follows. In Section \ref{sec:Preliminaries}, concepts regarding iris classification and SNN are introduced. Section \ref{sec:proposal} presents the methodology used to solve the problem, meanwhile, in Section \ref{sec:experimentFramework}, the dataset used is presented, as well as the pre-processing operations and the GSNN architecture. Then, experiments, results and discussion are outlined in Section \ref{sec:settings} and finally, Section \ref{sec:conclusions}, provides conclusions and guidelines for future work.


\section{Preliminaries}\label{sec:Preliminaries}
In this section, concept related to iris classification and Siamese Neural Networks (SNN) are introduced. More specifically, in Section \ref{sec:MLbiometrics} iris classification steps are described, whereas in Section \ref{sec:snn} Siamese Neural Networks are introduced.

\subsection{Iris classification}\label{sec:MLbiometrics}
\begin{figure}[!htbp]
  \centering
    \includegraphics[width=0.4\linewidth]{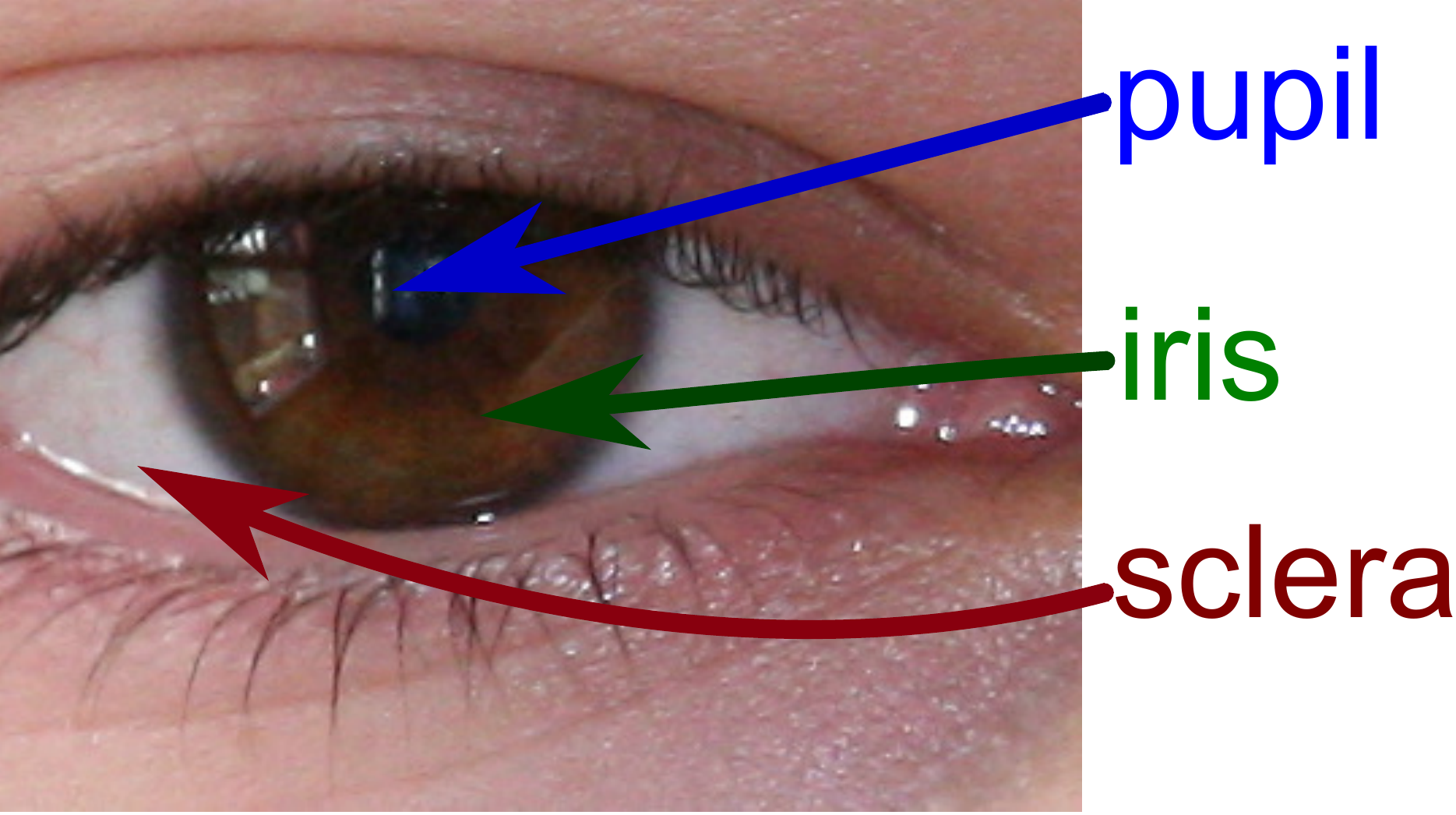}
    \caption{Image of iris, pupil and sclera in human eye.}
    \label{fig:eyes}
\end{figure}

Traditionally, the biometric iris recognition process involved $5$ key steps \cite{nguyen2017long, choudhary2012survey}.
In step 1, the subject's iris (or irises) is acquired using (various) cameras and optical sensors. Of course, in this phase, it should be ensured that the entire eye is captured, i.e., at least the pupil, iris, and sclera (Figure \ref{fig:eyes}). Then, in step 2, several pre-processing operations can be performed to reduce eventual noise, enhance the quality of the images, reduce dimensionality, and so on. In this step, an operation of segmentation is also performed, which usually consists in isolating the region of interest (i.e., the iris portion). This operation strongly depends on the image quality. In step 3, the iris regions are normalized in order to fit the same (constant) dimensions. In fact, rotating the head, pupil dilatation or contraction, and so on, can generate different regions of interest. In step 4, highly discriminative features are extracted from each normalized image. Finally, in step 5, these features are used for training a classifier, which can be further used for classifying incoming subjects.

Since deep learning-based models have increasingly been used in different recognition systems in recent years, their good results have led researchers to apply them also in the biometric domain. More precisely, they are widely used in steps 4 and 5, i.e., for feature extraction and matching. For example, in \cite{minaee2019deepiris} and \cite{al2018multi}, convolutional neural networks (CNN) are used for extracting the feature vector to be used for the final classification. In \cite{choudhary2019approach}, a deep convolutional network and a Support Vector Machine (SVM) classifier are combined for detecting contact lenses in human eyes using high-quality raw iris images, whereas in \cite{yan2018hierarchical}, authors propose a liveness detection based on Hierarchical Multi-class Iris Classification (HMC). In \cite{dua2019biometric}, a method based on a feed-forward architecture and k-means clustering algorithm for iris pattern classification is introduced, whereas, in \cite{jayanthi2021effective}, authors propose an approach in which the Hough Transformation is applied to localize the region of interest (i.e., iris region), and then a CNN is used for the classification. Another well-explored scope in iris biometrics involves the gender-identification. In particular in \cite{khalifa2019deep} deep convolutional neural are applied for this goal, while in \cite{khan2021authentication}, invariant moments and SVM are combined. Super-resolution convolutional neural networks (SRCNNs) for increasing the iris resolution from selfie images and the final gender classification are used in \cite{tapia2019sex}. Despite all these approaches, however, other work suggests that the gender-related information in the iris is not enough, and they advise using the whole periocular region \cite{kuehlkamp2019predicting}.


\subsection{Siamese Neural Networks (SNNs)}\label{sec:snn}
\begin{figure}[!htbp]
  \centering
    \includegraphics[width=0.6\linewidth]{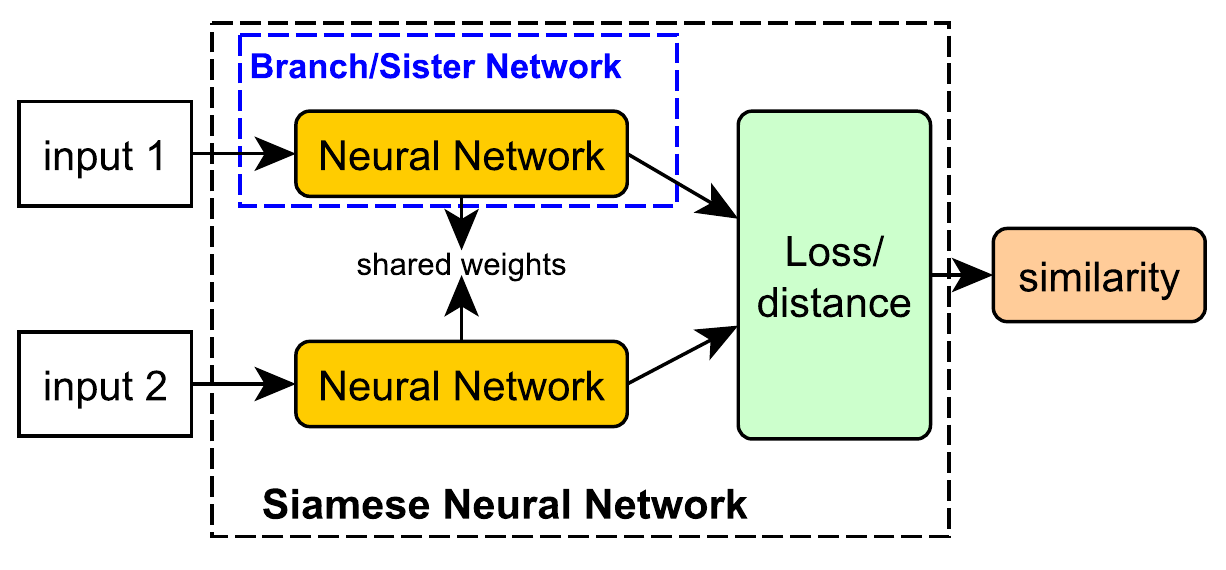}
    \caption{General architecture of a Siamese Neural Network (SNN).}
    \label{fig:siamese}
\end{figure}

Siamese Neural Networks (SNNs) are a type of neural network composed of multiple instances of the same model \cite{bromley1993signature}, which share the same architecture and weights, but different inputs called \textit{branch or sister networks} as shown in Figure \ref{fig:siamese}. The outputs of these two branch networks are combined in an additional layer (loss/distance layer), which is in charge of detecting if both outputs belong to the same input or not. This architecture is well-suited for measuring the similarity among the network inputs. In fact, the idea of the model is to try to minimize the loss (or distance) between inputs of the same class while trying to maximize the loss (or distance) between inputs of different classes \cite{chung2017two}. 

One of the main advantages of using SNNs is that if a new class is introduced in the dataset, it is unnecessary to retrain the whole model. This is because such networks do not learn how to predict specific classes, but they learn how to measure the similarity between two inputs.

SNNs have shown their usefulness in facial expression recognition \cite{hayale2019facial}, face recognition \cite{wang2014face}, matching local image patches \cite{hanif2019patch}, and generic image retrieval \cite{melekhov2016siamese}, and for object tracking \cite{guo2017learning}. Furthermore, these networks have shown an advantage when used with limited data \cite{chicco2021siamese}. SNN has also been used in biometric tasks, for example, for identifying if pairs of speech samples belong to a ``genuine" or ``replayed" user \cite{sriskandaraja2018deep}. In \cite{giot2019siamese}, SNNs are used to compare the similarity between two keystroke patterns and so lead the authentication process. A similar approach is applied in \cite{zhong2018palmprint}, however in this case for evaluating the similarity of two input palmprints according to their convolutional features.

To the best of our knowledge, SNNs are not yet explored for iris verification operations, as well as their applications with graphs directly extracted from LR iris images. For this reason, this work represents a first step toward the usage of LR iris images for extracting graph-based information and then applying GSNN technology directly over this graph structure as a verification system.

\section{Methodology}\label{sec:proposal}
In this work, we propose a novel methodology for LR iris verification, i.e., to check if two iris images belong or not to the same user. To do so, our approach is based on graph analysis and Graph Siamese Neural Networks (GSNN) and it is divided into two phases. In the first one, we preprocess the LR iris images and convert them into graph-based structures (Section \ref{sec:extraction}). Then, in the second phase, we use graph information to train a GSNN able to evaluate graphs’ similarities and predict if they belong to the same subject or two different ones (Section \ref{sec:gsnn}).

\subsection{Iris to Graph}\label{sec:extraction}

In the process of converting the images to graphs, our approach proposes to study and use two different types of images: the original LR iris images or a version of these images in which spectral components are highlighted. These spectral components are obtained by using a convolutional filter on the original image, before carrying out the steps to convert the image into a graph. In this way, it is possible to obtain more sharped images, favouring the extraction of more populated graphs. More specifically, in this work, the untouched LR iris images are called \textit{original} images, whereas, the images in which convolution filters are applied, are called \textit{spectral enhanced} images.

\begin{figure}[!htbp]
  \centering
    \includegraphics[width=0.7\linewidth]{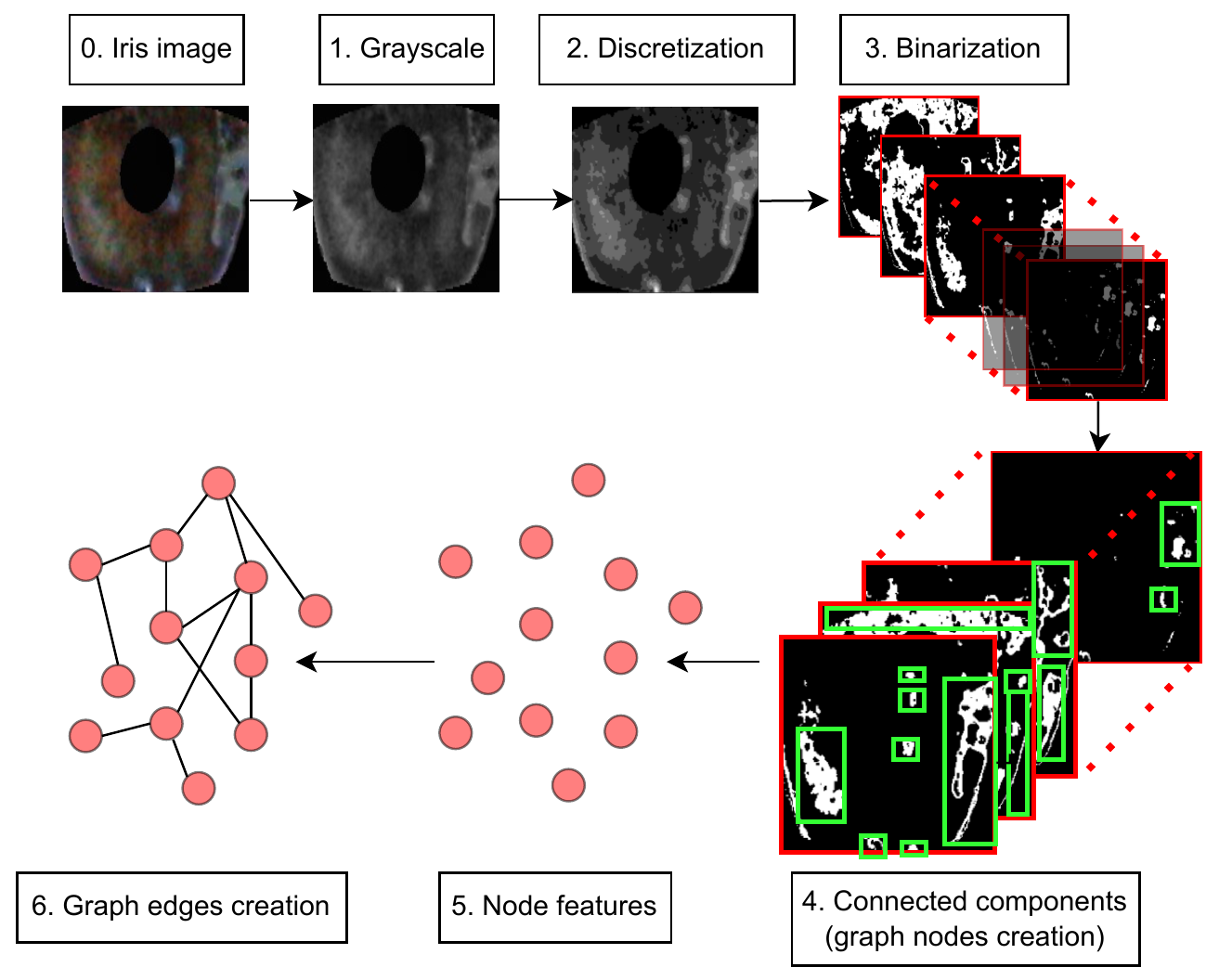}
    \caption{Methodology for extracting graph structure from iris images.}
    \label{fig:methodology}
\end{figure}

For both types of images, the steps to extract a graph structure are the same. In particular, as shown in Figure \ref{fig:methodology}, $6$ different operations are performed:

\begin{enumerate}

\item \textit{Grayscale:} the input image is converted to grayscale.

\item \textit{Discretization:} the pixels in the grayscaled image have intensity values in a range from 0 to 255. During the discretization phase, they are grouped into sectors (or bins) of a fixed size ($\delta$). In this sense, if a pixel is within a specific range, the pixel takes the value of the minimum value of the bin. This means that the pixels in the image can only have $B = \lceil$255/$\delta \rceil$ possible values. This approach helps us in limiting the possible values that each pixel can take, homogenizing similar areas in the image. This new discretized image is indicated as $\Delta G$. More specifically, in our analysis the $\delta$ is set to 20, generating 13 different bins.  

\item \textit{Binarization:} a binary image is created for each bin with the pixels corresponding to that bin set to 1 and all the others to 0. This operation is repeated for all the available bins, excluding the first bin, where the intensity value is 0.  In this way, a single $\Delta G$ generates $B$-1 different binary images. In this work, 12 different binary images are obtained from each $\Delta G$. Through this binarization, the idea is to highlight connected components that form homogeneous areas in the image.

\item \textit{Graph node creation:} in each of the binary images, connected components, i.e., areas where several consecutive pixels are 1, are detected using an 8-way (default) connectivity. All these components are identified using bounding boxes (bboxes), i.e., a minimum area rectangle around the given object that serves as the region of interest. All the bboxes detected in all the binary images of a certain $\Delta G$, will represent the nodes of the corresponding graph. A single binary image can contain multiple connected components.

\item \textit{Node feature:} for each detected node, i.e., for each bbox in the binary images, $F$ characteristics are extracted and used as a feature vector of the node. More specifically, in this work, 7 different features are extracted: the ratio of the component size with respect to the original image size (number of pixels of the component divided by the total number of pixels), the ratio of the component size with respect to the bbox that contains the component itself, the coordinate of the centre of the bbox, its height and width, and the number of connected components in the same bin.

\item \textit{Graph edge creation:} to create the edges in the graph, the positions of the bboxes in the binary images are analyzed. In particular, if two components share the same position in the image, either in the same bin or in the other bins that belonged to the same $\Delta G$, these nodes are connected through an edge. The weight of this edge is then obtained by calculating the Euclidean distance between the centres of the bboxes that contain the components separately.
\end{enumerate}

\subsection{Verification system}\label{sec:gsnn}
Once the (\textit{original} or \textit{spectral enhanced}) images are converted into graphs, it is possible to use them for training the verification system based on GSNN. A GSNN is an SNN that is able to directly analyze graph structures, i.e., it receives as input a pair of adjacency and feature matrices representing the two graphs to be compared and returns a similarity value between them.
However, in order to be used as inputs for the GSNN and be correctly compared, all the graphs in the dataset must have the same number of nodes (N), and so have adjacency matrices of $N \times N$. At the same time, the feature matrix, i.e., a matrix in which the $i-th$ row contains the feature vector of the $i-th$ node, should be $N \times F$. In fact, GSNNs are composed of at least one graph convolutional (GCN) layer that is able to process both the adjacency and the feature matrices.

In this scenario, using the iris to graph methodology introduced in Section \ref{sec:extraction}, all the graphs are generated with a different number of $N$. For this reason, we use \textit{padding} for filling the adjacency and feature matrices of the graphs composed of less than $N$ nodes. This padding is performed by adding rows of zeros to the feature matrices and rows and columns of zeros to the adjacency matrices. Once all the matrices that describe the graphs have the same dimensionality, they can be paired and used for training and evaluating the GSNN.

\section{Experimental Framework}\label{sec:experimentFramework}
In this section, the dataset, pre-processing operations, and models are described. In particular, in Section \ref{sec:dataset}, the used dataset, as well as the operations used for pre-processing LR iris images are described. Finally, the GSNN architecture is introduced in Section \ref{sec:models}.

\subsection{Dataset and Pre-processing}\label{sec:dataset}
\begin{figure}[]
\centering
  \begin{subfigure}{.22\textwidth}
    \includegraphics[width=\linewidth]{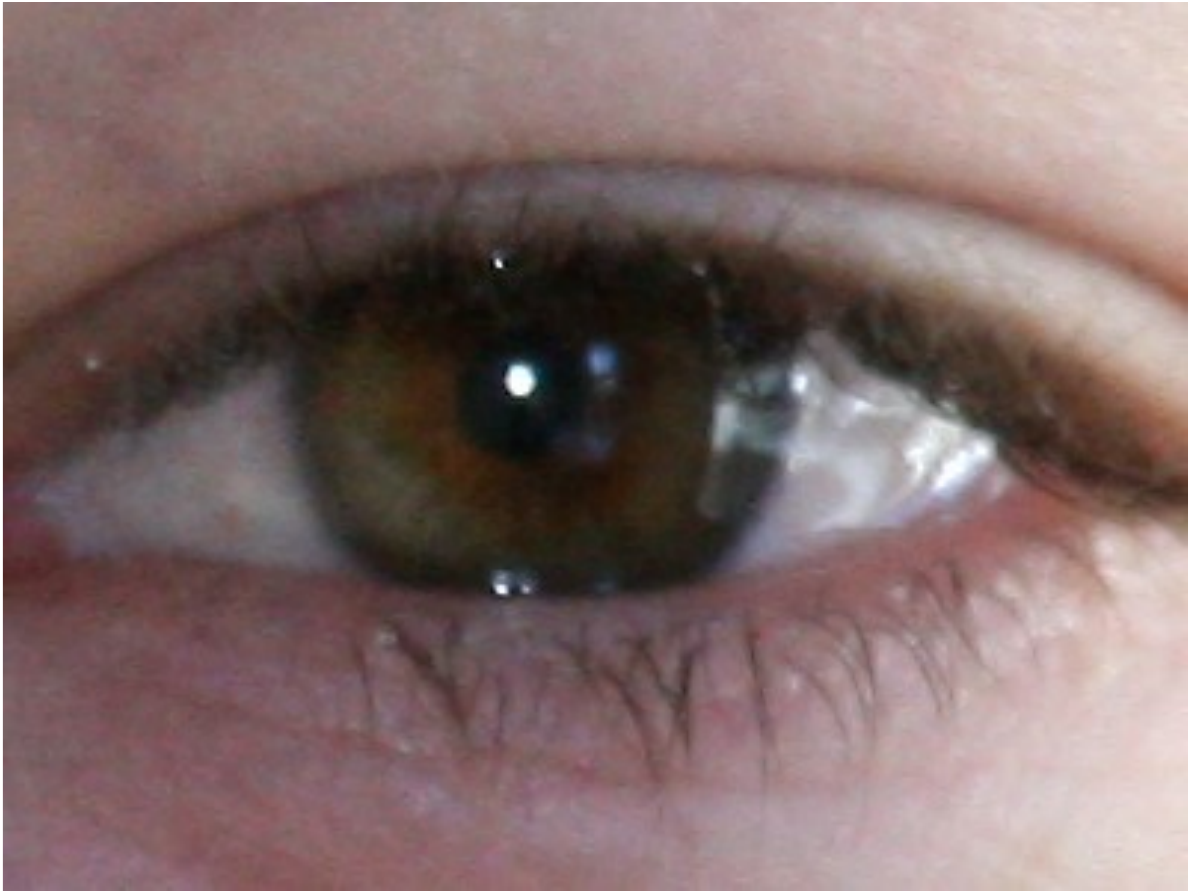}
  \subcaption{Original iris image (UBIRIS-V2)}
  \label{fig:ubiris}
\end{subfigure}
  \begin{subfigure}{.22\textwidth}
  \centering
    \includegraphics[width=\linewidth]{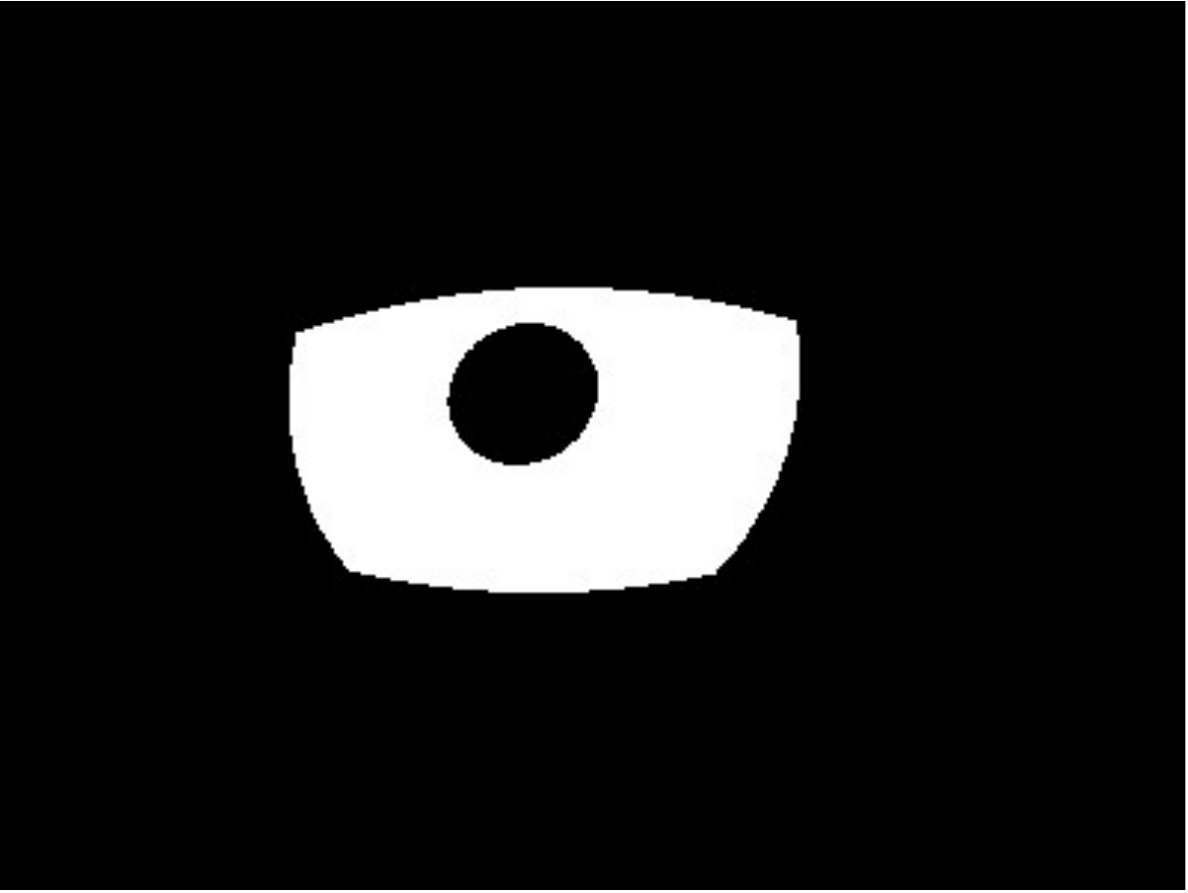}
  \subcaption{Iris masked image (IRISSEG-EP)}
  \label{fig:mask}
\end{subfigure}

  \begin{subfigure}{.15\textwidth}
    \includegraphics[width=\linewidth]{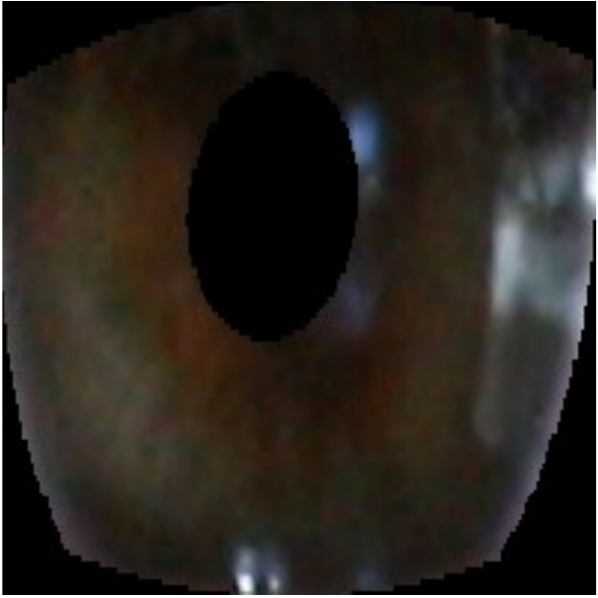}
  \subcaption{Cropped and re-scaled image}
  \label{fig:cropped_resize}
\end{subfigure}
\begin{subfigure}{.15\textwidth}
    \includegraphics[width=\linewidth]{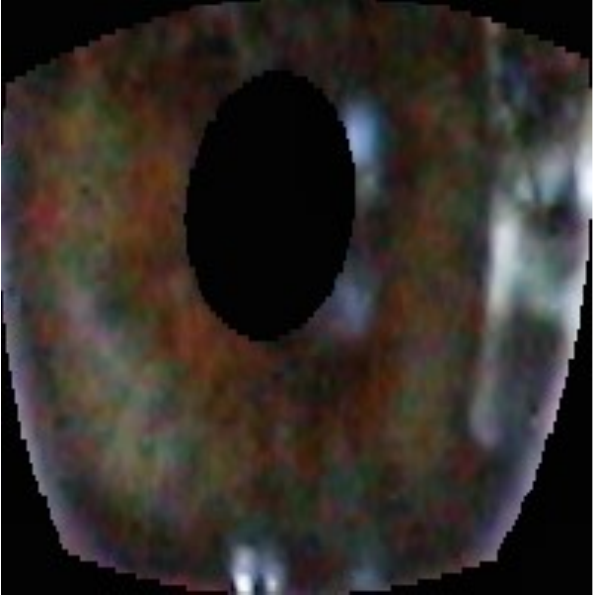}
  \subcaption{Histogram equalization}
  \label{fig:stretching}
\end{subfigure}
\begin{subfigure}{.15\textwidth}
    \includegraphics[width=\linewidth]{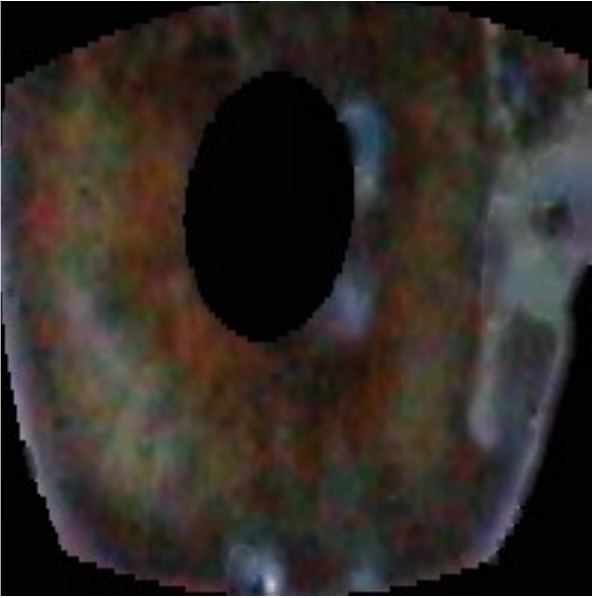}
  \subcaption{Reflection artefacts removed}
  \label{fig:reflection}
\end{subfigure}
    \caption{Image pre-processing steps.}
    \label{fig:dataset_preprocessing}
\end{figure}

In this work, a dataset called UBIRIS-V2 \cite{proencca2009ubiris} was used. The UBIRIS-V2 contains iris images captured under noisy conditions (at-a-distance, on the move, and on the visible wavelength). The dataset contains iris images of $261$ different persons, for each of whom $15$ pictures per iris were taken at each meter between $4$ and $8$ m, away from the camera, i.e., $3$ pictures at each distance. Each subject participates in two shooting sessions, in each of whom was required to walk slower than normal speed while looking at specific markers. The UBIRIS-V2 database is distributed as $11,102$ single iris images, as indicated in Table \ref{tab:dataset_stat}. Each image is cropped to a shape of 400 x 300 pixels (Figure \ref{fig:ubiris}).

This dataset was combined with the IRISSEG-EP dataset \cite{hofbauer2014ground}. This dataset includes $2,250$ iris masks of $100$ users of the UBIRIS-V2, as shown in Figure \ref{fig:mask}. The IRISSEG-EP dataset allows highlighting only the iris part of the images, removing all the areas that can introduce noisy information such as eye-brown, skin, pupil, and sclera. The masked iris images were then cropped using the minimum rectangle able to contain just the highlighted area. However, this cropping approach generates images of different sizes, according to the iris area in each image. For this reason, in order to homogenize all the cropped images, we re-scaled them to a fixed dimension of 200x200 pixels (Figure \ref{fig:cropped_resize}).

\begin{table}[]
\centering
\begin{tabular}{ccc}
\hline
  & \multicolumn{1}{c}{\begin{tabular}[c]{@{}c@{}}UBIRIS-V2\end{tabular}} & \multicolumn{1}{c}{\begin{tabular}[c]{@{}c@{}}IRISSEG-EP\end{tabular}} \\\hline
  \textit{type} &iris &mask \\[2mm]
    \textit{\# subject} &261 &100  \\ [2mm]
    \textit{\# irises} &522 &- \\ [2mm]
    \textit{\# images} &11,102 &2,250  \\ [2mm]
    \textit{image size} &400x300 &400x300\\ [2mm]
    \textit{distance (m)} &4, 5, 6, 7, 8 &- \\ \hline
\end{tabular}
\caption{Overview about the used datasets}
\label{tab:dataset_stat}
\end{table}

Then, an operation of histogram equalization is performed in all the cropped iris images, as shown in Figure \ref{fig:stretching}. This equalization (histogram stretching) allows to reshape the intensity of all the pixels, i.e., adapt the pixel distribution in the full range of possible status \cite{kaur2011survey}. In this way, the contrast of the image is enhanced. Finally, an operation to remove undesired reflections from the image was performed. These reflection artefacts are white-dot areas that can introduce noisy information and reduce the efficiency of visual computing system \cite{wan2017benchmarking}. In particular, following the indications provided in \cite{jamaludin2016removal}, the boundaries of these artefacts were detected, and then an interpolation operation was performed in order to fill the white area (Figure \ref{fig:reflection}).

These pre-processed iris images (2,250) of 100 different users represent our initial dataset, and they are used as the inputs for our analysis. From now on, the terms \textit{iris dataset} is used for referring to this enhanced dataset.

\subsection{GSNN Architecture}\label{sec:models}
In this work, we use a model able to process graph information as a branch network for the Siamese Neural Network, creating the GSNN. More specifically, each branch network was composed of two GCN layers with 500 and 200 neurons, respectively. Each GCN had a \textit{tanh} activation function. The two GCNs layers were followed by a global average pooling \cite{lin2013network}, before generating the output through a final dense layer of $200$ neurons and ReLu activation. 

The outputs of the branch networks (embedding vectors) were finally combined in order to compute the similarity among their respective inputs. More specifically, the two embedding vectors were subtracted between them, and used as input to a Deep Neural Network (DNN) composed of three hidden layers with 400, 200, and 100 neurons, respectively.

\section{Experimental study}\label{sec:settings}
In this Section, experiments and results are introduced and further discussed. More specifically, in Section \ref{sec:experiments} the experiments carried out are presented, whereas in Section \ref{sec:results} the results obtained are presented. Finally, in Section \ref{sec:discussion}, a discussion about the benefits and limitations of the introduced approach is reported.

\subsection{Experiments}\label{sec:experiments}

\textbf{Experiments 1.}
In the first experiment, we use all the \textit{original} images of 10 randomly selected users from our iris dataset. For each user, the images are firstly split into the train, validation, and test dataset using a proportion of 60\%, 20\%, and 20\%, respectively. In this way, all the users appear in all the datasets, although with images taken at different distances. Then, in each dataset, the images are converted into graphs using the methodology introduced in Section \ref{sec:extraction}. Furthermore, as introduced in Section \ref{sec:gsnn}, an operation of padding is applied in order to homogenize graph population, i.e., obtain graphs with the same dimensionality in both adjacency and feature matrices. 

In this experiment, five different numbers of nodes ($N$) have been used for evaluating how these parameters affect the classification performance. More specifically, 100, 150, 200, 250, and 300 nodes are used. In this sense, if a graph is characterized by a major number of nodes of the selected one, it is excluded from the experiment. 

Finally, when the image datasets (train, validation, and test) are converted into graph datasets, an operation for creating the pairs to be used in the GSNN training is performed. In particular, each graph in the dataset is paired with all the graphs of the same user and with as many graphs randomly selected from the others users.

\textbf{Experiments 2.}
In the second experiment, we study how the spectral component of the image affects our graph extraction methodology and so the final GSNN performance. For this purpose, we start using the \textit{spectral enhanced} images of 10 randomly selected users from our iris dataset. These images are generated using the \textit{original} images and applying a convolution operation to highlight their spectral components, as introduced in Section \ref{sec:extraction}. More specifically, the convolution filter shown in Table \ref{tab:filter} is used, testing separately 6 different values of $\alpha$. Then, as for experiment 1, for each user, the spectral images are split into train, validation, and test datasets using a proportion of 60\%, 20\%, and 20\%, respectively. Again, five different sizes for the number of nodes ($N$) are used for evaluating the classification performance, and the paired datasets are created.

\begin{table}
\begin{tabular}{ ccccccc }
  \begin{tabular}[c]{@{}c@{}}filter\\ structure \end{tabular} & \begin{tabular}[c]{@{}c@{}}filter 1\\ ($\alpha$) \end{tabular} & \begin{tabular}[c]{@{}c@{}}filter 2\\ ($\alpha$) \end{tabular} & \begin{tabular}[c]{@{}c@{}}filter 3\\ ($\alpha$) \end{tabular} & \begin{tabular}[c]{@{}c@{}}filter 4\\ ($\alpha$) \end{tabular} & \begin{tabular}[c]{@{}c@{}}filter 5\\ ($\alpha$) \end{tabular} & \begin{tabular}[c]{@{}c@{}}filter 6\\ ($\alpha$) \end{tabular} \\\hline 
 $\begin{bmatrix}
0 & \alpha & 0\\
\alpha & 1 & \alpha\\
0 & \alpha & 0
\end{bmatrix}$ & 1/10 & 1/9 & 1/8 & 1/7 & 1/6 & 1/5 \\  
\end{tabular}

\caption{Convolutional filters used in the experiment 2.}
\label{tab:filter}
\end{table}

\textbf{Experiments 3.}
In the third experiment, the idea is to evaluate how the performance of the GSNN changes when the number of users increases, i.e., evaluate the generalization abilities of our approach. For this reason, we start by selecting the two best configurations obtained in experiment 2, and we repeat their implementations, i.e., re-train each time a new model, using a different number of initially selected users. More specifically, 10, 20, 30, 40, and 50 different users have been used.

\subsection{Results}\label{sec:results}

\textbf{Experiments 1.}
As shown in Table \ref{tab:first}, selecting all the images of 10 different users, the iris dataset is composed of 299 images, generating as many as graphs. The table shows that with a low number of nodes for all the graphs (100 nodes), just 19 graphs are removed, i.e., about 6\% of the dataset. On the other hand, by increasing the number of nodes to 150, just 2 graphs are removed, and no one for the other considered sizes. 

In terms of model performance, although including graphs with a higher number of nodes, increase the GSNN classification abilities, it happens until reaching a specific size (200), after that the performance starts to decrease. This trend is highlighted in Table \ref{tab:first}, where the GSNN that uses graphs with 100 nodes shows accuracy and F1-score of 58.00\% and 66.68\%, respectively, whereas using graphs with 200 nodes, both metrics increase their values of $\sim$3\% and $\sim$6\%. However, when graphs with 250 and 300 nodes are used, the model performance gets worse. The same trend is also confirmed in the test results.


\begin{table}[]
\begin{tabular}{c|cc|cc|cc}
\textbf{\# nodes} & \textbf{\begin{tabular}[c]{@{}c@{}}\# graphs\end{tabular}} & \textbf{\begin{tabular}[c]{@{}c@{}}\# graphs\end{tabular}}  & \multicolumn{2}{c|}{\textbf{Validation}} & \multicolumn{2}{c}{\textbf{Test}} \\ 

& \textbf{removed} & \textbf{used} & \textbf{Accuracy \%} & \textbf{F1-score \%} & \textbf{Accuracy \%}  &\textbf{F1-score \%}
\\\hline
100 & 19 & 280 & 58.00 &66.68 & 58.63 &68.51\\
150 & 2 & 297 & 61.17 &69.06 & 58.86 &69.37\\
200 & 0 & 299 & 61.28 &72.85 & 65.57 &77.20\\
250 & 0 & 299 & 62.00 &71.46 & 62.58 &73.80\\
300 & 0 & 299 & 60.72 &67.69 & 59.57 &68.74\\

\end{tabular}
\caption{Validation and test accuracy and F1-score in experiment 1.}
\label{tab:first}
\end{table}

\begin{figure}[!htbp]
  \centering
    \includegraphics[width=0.6\linewidth]{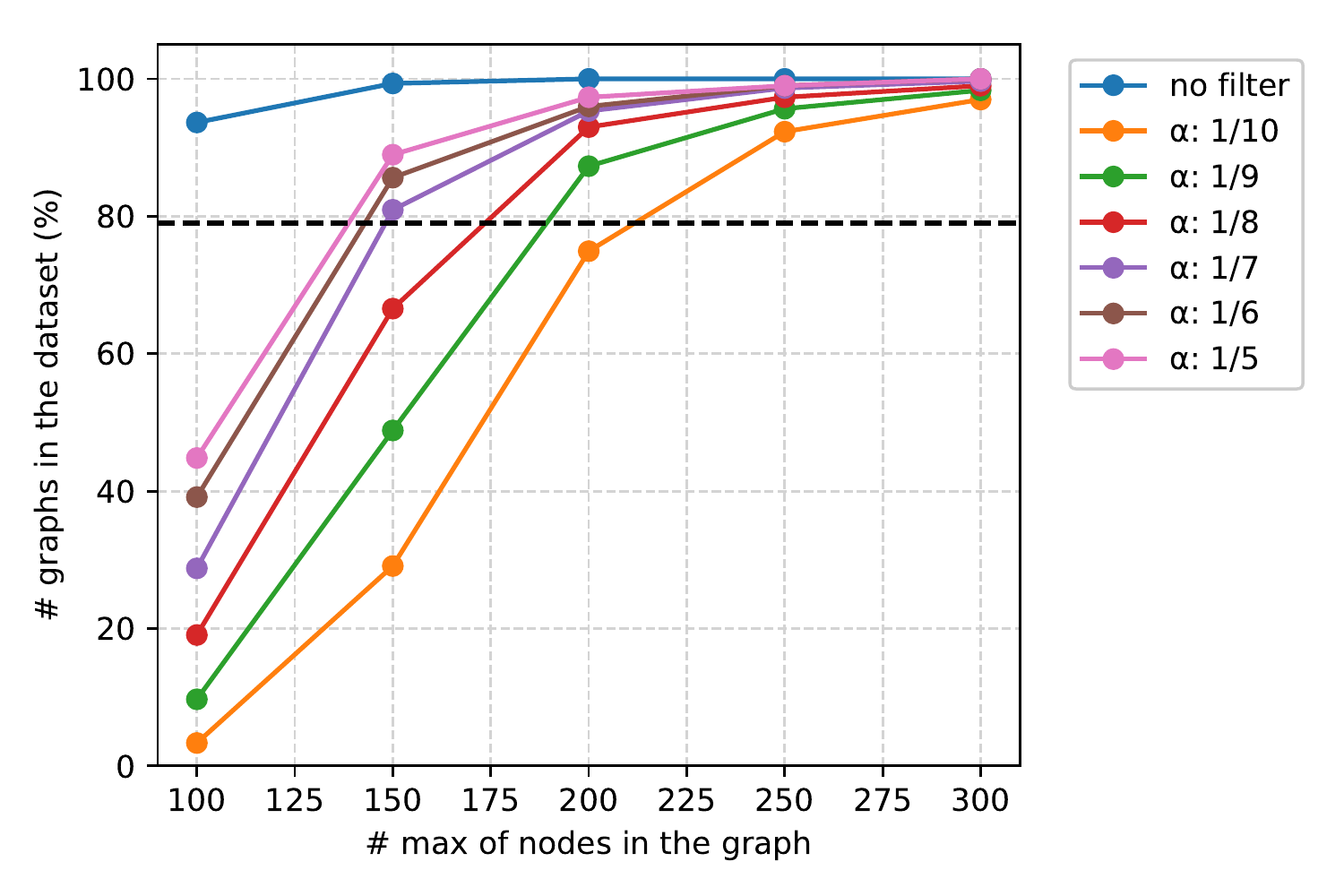}
    \caption{Number of graphs (in percentage) used in experiment 2, depending on the selected number of nodes and the used convolutional filter.}
    \label{fig:graph_second}
\end{figure}

\textbf{Experiments 2.}
Using a different convolutional filter over the image dataset allows one to highlight the spectral component of the images. This spectral enhancement promotes the generation of graphs with a higher number of nodes. This trend is confirmed in Figure \ref{fig:graph_second}, in which, it is possible to appreciate that, for a fixed number of nodes, the number of removed graphs increases by decreasing the $\alpha$ parameter in the used convolutional filter. For example, as shown in Figure \ref{fig:graph_second}, about 95\% of the graphs extracted from no filtered images have less than 100 nodes, whereas using $\alpha$ = 1/5, only about 42\% of the graphs have less than 100 nodes, and using $\alpha$ = 1/10 almost all the graphs have more than 100 nodes, and so they are removed.

Following the distributions reported in Figure \ref{fig:graph_second}, only the configurations in which at least the 80\% of graphs are available (above the dotted line), are used in experiment 2. This is because a huge reduction of the graphs dataset can skew the results and so the evaluation of our approach. In particular, Table \ref{tab:second} reports the performance of all the configurations in terms of Accuracy and F1-score. The table shows that the highest accuracy (65\%) is obtained using $\alpha$ = 1/7 and limiting the analysis only over graphs with 150 nodes, which represents about 81\% of the input population. A similar accuracy value (64.87\%) is also obtained using $\alpha$ = 1/9 and graphs composed of no more than 250 nodes. On the other hand, in terms of F1-score, the best solutions are obtained by using $\alpha$ = 1/7 and $\alpha$ = 1/5 and using graphs with no more than 200 and 250 nodes, respectively. More specifically, in these two configurations F1-score values $\geq$ 77\% are obtained. For this reason, these two configurations are the ones used in the third experiment.

\begin{table}[]
\centering
\begin{tabular}{l|ccccc|ccccc}
& \multicolumn{5}{c|}{\textbf{Validation Accuracy \%}} & \multicolumn{5}{c}{\textbf{Validation F1-score \%}} \ \\\hline
 \textbf{filter}  & \textbf{100} & \textbf{150} & \textbf{200} & \textbf{250} & \textbf{300} & \textbf{100} & \textbf{150} & \textbf{200} & \textbf{250} & \textbf{300} \\ \hline
\textbf{no filter} &58.00 &61.17 &61.28 &62.00 &60.72 &66.68 &69.06 &72.85 &71.46 &67.69 \\
\textbf{$\alpha$ : 1/10} & - & - & - & 60.96 & 63.50 &- &- &- &68.74 &72.25\\
\textbf{$\alpha$ : 1/9} & - & - & 63.20 & \textbf{64.87} & 61.25 &- &- &73.78 &74.61 &68.82\\
\textbf{$\alpha$ : 1/8} & - & - & 57.14 & 61.62 & 60.96 &- &- &63.36 &72.67 &74.50\\
\textbf{$\alpha$ : 1/7} & - & \textbf{65.00} & 63.48 & 60.54 &59.96 &- &74.81 &\textbf{77.22} &69.78 &69.68\\
\textbf{$\alpha$ : 1/6} & - & 62.60 & 61.71 & 56.75 &62.00 &- &73.22 &70.28 &60.95 &74.00\\
\textbf{$\alpha$ : 1/5} & - & 62.10 & 61.75 & 62.92 & 61.16 &- &70.13 &69.74 &\textbf{77.18} &68.25
\end{tabular}
\caption{Validation Accuracy and F1-score in experiment 2. Top values are in bold.}
\label{tab:second}
\end{table}

\textbf{Experiments 3.}
Table \ref{tab:thid} reports the performance of the best configurations detected in the second experiment when the number of different input users is increased. Table \ref{tab:thid} shows that increasing the number of users involved in the verification process has a different impact on the overall population. In fact, in the first configuration ($\alpha$=1/7 and 200 nodes) increasing the number of users, the population decreases from 95.32\% to 92.50\%, whereas in the second configuration ($\alpha$=1/5 and 250 nodes), although varying the number of users, the population is quite stable ($\sim$ 99\%).
On the other hand, the results show that by increasing the number of users, the performance of the GSNN worsens, especially in terms of F1-score. This trend is confirmed in the first configuration in both validation and test metrics, starting from an F1-score $\geq$77\% with just 10 users, and reaching a value of $\leq$75 when 50 users are involved. The same trend is also confirmed in the validation performance (F1-score) when the second configuration is used.

\begin{table}[]
\resizebox{\linewidth}{!}{
\begin{tabular}{c|c|cc|cc|c|cc|cc}
\textbf{} & \multicolumn{5}{c|}{\textbf{\begin{tabular}[c]{@{}c@{}}Configuration 1\\ \textbf{$\alpha$: 1/7} - 200 nodes\end{tabular}}} & \multicolumn{5}{c}{\textbf{\begin{tabular}[c]{@{}c@{}}Configuration 2\\ \textbf{$\alpha$: 1/5} - 250 nodes\end{tabular}}} \\\hline
\textbf{\#} &\textbf{used} & \multicolumn{2}{c|}{\textbf{Validation}} & \multicolumn{2}{c|}{\textbf{Test}} & \textbf{used} & \multicolumn{2}{c|}{\textbf{Validation}} & \multicolumn{2}{c}{\textbf{Test}} \\
\textbf{users} &\textbf{graphs \%} & \textbf{Accuracy \%} & \textbf{F1-score \%} & \textbf{Accuracy \%} & \textbf{F1-score \%} & \textbf{graphs \%} & \textbf{Accuracy \%} & \textbf{F1-score \%} & \textbf{Accuracy \%} & \textbf{F1-score \%}\\\hline
\textbf{10} & 95.32 &63.48 &77.22 & 65.55 & 78.62 &99.00 &62.92 &77.18 &66.15 &79.52 \\
\textbf{20} & 93.50 & 63.76 & 76.61 & 66.00 &78.74 &98.83 &63.16 &75.20 &64.72 &77.08\\
\textbf{30} & 93.31 & 63.46 & 75.84 & 65.81 & 77.65 &98.89 &61.83 &75.89 &65.87 &78.90 \\
\textbf{40} & 93.65 & 64.85 & 75.99 & 65.97 & 77.41 &99.08 &62.76 &74.51 &62.76 &75.06\\
\textbf{50} & 92.50 & 64.24 & 74.64 & 66.12 & 76.91 &98.80 &61.83 &74.23 &66.06 &77.87
\end{tabular}}
\caption{Results obtained in experiment 3.}
\label{tab:thid}
\end{table}

\subsection{Discussion}\label{sec:discussion}
The introduced approach shows how to use graph representation of LR iris image for verification task. In this sense, our approach shows that, although spectral images, i.e., images in which components are highlighted using convolutional filters, allow one to improve the accuracy in the verification tasks, they also generate more populated graphs. Hence, when we use a fixed number of nodes, most of them are removed during the execution and these changes in the population can be one of the reasons for the obtained improvements, if the most difficult examples are removed. Anyway, if this is true, one could also use this technique to reject images that could not be correctly classified.

Table \ref{tab:comparison} shows information about state-of-the-art works that use the same LR UBIRIS-V2 dataset for different biometrics tasks. As it is possible to see from the Table, the majority of the studies are focused on the segmentation of these images, i.e., in detecting and highlighting iris (or sclera) from these LR images. On the other hand, just a few works are focused on LR iris recognition, and these works achieved higher accuracy values than the ones reached by our approach. For this reason, although this work shows promising results, they can be further improved by investigating image preprocessing operations (histogram equalization and removing reflection artefacts) which are out of the scope of this paper.

\begin{table}[]
\begin{tabular}{cccc}
\hline
\textbf{Target} &\textbf{Task}  &\textbf{Classifier} & \textbf{\begin{tabular}[c]{@{}c@{}}Accuracy\\\%\end{tabular}} \\ \hline
sclera \cite{lucio2018fully} &segmentation   &- &- \\
iris \cite{jha2020pixisegnet} &segmentation  &- &- \\
iris \cite{lian2018attention} &segmentation  &- &- \\
iris \cite{tan2011automated} &segmentation  &- &- \\
iris \cite{liu2016accurate} &segmentation  &- &- \\
iris \cite{kaur2018iris} &recognition  &k-NN &$\sim$ 94 \\
periocular \cite{proencca2017deep} &recognition &CNN &$\sim$ 88 \\\hline

\end{tabular}
\caption{Comparison among state-of-the-art works that uses the LR UBIRIS-V2 dataset.}
\label{tab:comparison}
\end{table}

\section{Conclusions and Future work}\label{sec:conclusions}


In this work, a novel approach for long-range (LR) iris verification has been proposed. More specifically, we used a novel approach to extract graph information from each LR iris image captured from standard cameras, and then use these graphs for training a GSNN, with the aim to detect whether two iris belong to the same person or not. 

Three experiments have been carried out to measure the robustness, solvency and generalization performance of the proposed approach. In the first one, we analyzed how the number of nodes in the graphs affects the verification process. In particular, results show that a small number of nodes or using too many nodes worsens the model performance. In this sense, it is important to find a good trade-off. In the second experiment, we analyzed how the spectral component of the images affects the graph definition and so the verification task. In this sense, we applied a convolutional filter to the image before the operation of graph extraction. This approach shows that it is possible to improve the accuracy and F1-score of the GSNN using a specific filter. However, at the same time, enhancing the spectral information, larger graphs are extracted, which causes some of them to be excluded from the analysis. Finally, in the third experiment, we evaluated how the number of users impacts the verification system performance. In this sense, the results show that increasing the number of users from 10 users to 50, it slightly affects GSNNs performance (small deterioration).

This study demonstrates the possibility of working with graphs to analyze biometric information, since the results obtained are promising for further work in this line. Results above 78\% in terms of F1-score, in a task as complex as knowing how to detect when an iris belongs to the same person in a poorly controlled context, can encourage the community to continue advancing in this direction. However, it is obvious that in a user identification system or in critical applications, this level of accuracy is not sufficient and could not be used safely as the sole method of identification. Therefore, for the time being, LR iris recognition technology can only be used to support and enhance the security of other verification systems in a controlled environment (as a second authentication factor, for example).

As already pointed out in Section \ref{sec:discussion}, there is room for improvement to achieve better results by improving the iris preprocessing task. For example, applying image operations such as canny or Sobel filters to increase the contrast between the different areas of the iris. This process can improve the current graph generation pipeline allowing the integration of more information about connected components and their boundaries. On the other hand, the potential of the GSNN model can be tested with other types of biometric information, such as fingerprints or faces, for example. In this way, the procedure would be similar (transform the images into graphs and train the model with these graphs). Finally, instead of applying the solution to detect whether two images are of the iris of the same person, we could work on liveness detection, i.e., the model could be able to interpret whether the image presented is an iris image captured at the time of the person or whether it is an attempt at impersonation in which, for example, a previously obtained image of the iris is being used.

\begin{acks}
This work has been partially supported by the Spanish Centre for the Development of Industrial Technology (CDTI) under the project ÉGIDA (EXP 00122721 / CER-20191012) - RED DE EXCELENCIA EN TECNOLOGIAS DE SEGURIDAD Y PRIVACIDAD

\end{acks}

\bibliographystyle{ACM-Reference-Format}
\bibliography{sample-sigplan}


\end{document}